\documentclass[preprint,12pt]{elsarticle}

\usepackage{amssymb}
\usepackage{amsmath,amssymb}
\usepackage{lmodern}
\usepackage{iftex}
\usepackage{lineno}
\usepackage{textcomp} 
\usepackage{xcolor}
\usepackage{longtable,booktabs,array}
\usepackage{multirow}
\usepackage{calc} 

\usepackage{hyperref}       
\usepackage{url}            
\usepackage{booktabs}       
\usepackage{amsfonts}       
\usepackage{nicefrac}       
\usepackage{microtype}      
\usepackage{graphicx}
\usepackage{boldline}
\usepackage[ruled,vlined]{algorithm2e}
\usepackage[margin=1.0in]{geometry}

\journal{Pattern Recognition}
\begin{document}

\begin{frontmatter}

\title{Guided Frequency Loss for Image Restoration}

\author[label1]{Bilel Benjdira\corref{cor1}}
\ead{bbenjdira@psu.edu.sa}
\author[label1]{Anas M. Ali}
\ead{aaboessa@psu.edu.sa}
\author[label1]{Anis Koubaa}
\ead{akoubaa@psu.edu.sa}
\affiliation[label1]{organization={Robotics and Internet-of-Things Laboratory},
            addressline={Prince Sultan University}, 
            city={Riyadh},
            postcode={12435}, 
            country={Saudi Arabia}}

\cortext[cor1]{Corresponding author.}

\begin{abstract}
  Image Restoration has seen remarkable progress in recent years. Many generative models have been adapted to tackle the known restoration cases of images. However, the interest in benefiting from the frequency domain is not well explored despite its major factor in these particular cases of image synthesis. In this study, we propose the Guided Frequency Loss (GFL), which helps the model to learn in a balanced way the image's frequency content alongside the spatial content. It aggregates three major components that work in parallel to enhance learning efficiency; a Charbonnier component, a Laplacian Pyramid component, and a Gradual Frequency component. We tested GFL on the Super Resolution and the Denoising tasks. We used three different datasets and three different architectures for each of them. We found that the GFL loss improved the PSNR metric in most implemented experiments. Also, it improved the training of the Super Resolution models in both SwinIR and SRGAN. In addition, the utility of the GFL loss increased better on constrained data due to the less stochasticity in the high frequencies' components among samples.
\end{abstract}



\end{frontmatter}
\section{Introduction}

Image restoration is an umbrella term for many computer vision tasks that target improving the quality of a captured image \cite{ZHOU2023109602}\cite{SHEN2022108909}\cite{ZHAI2022108333}. It aims to exceed the limitation of the sensors and the capturing circumstances to synthesize a more realistic and vivid image \cite{LOPEZRUBIO20101835}\cite{WANG2022108867}\cite{XUE2023109041}. Like other domains affected by the rise of deep learning models, Image Restoration has evolved a lot in recent years \cite{HE2023109038}\cite{THAKUR2023109603}\cite{gu2022ntire}\cite{wang2022ntire}\cite{liang2021swinir}\cite{conde2023swin2sr}.  However, the increasing efficiency in this domain is sometimes inhibited by significant differences between restored and authentic images. These differences may be detected in the spatial domain, such as visible artifacts \cite{odena2016deconvolution} that appear while zooming the image. Some other differences between the restored and the authentic images can only be detected in the frequency domain \cite{jiang2021focal}.  For example, many studies \cite{wang2020cnn}\cite{zhang2019detecting}\cite{huang2020fakeretouch} have analyzed the frequency spectra of the generated images and found some periodic patterns that correlate with the visual artifacts in the spatial domain. The detected gap in the frequency domain may be due to the neural networks' spectral bias \cite{rahaman2019spectral}\cite{tancik2020fourier}.  This bias is inherited in Neural Networks, making them more able to learn the low frequencies better than the hard ones. Besides this phenomenon, other studies \cite{xu2019frequency} revealed the Frequency Principle of Deep Neural Networks (DNNs). This principle states that DNNs learn target functions from low to high frequencies during training. The spectral bias and the Frequency principle underline the hidden gap between the real and the restored images. Hence, enforcing DNN to learn the image's full spectral representation is challenging. 
In this study, a new loss function is introduced to narrow the aforementioned gap and to balance the learning of image restoration between the spatial and the frequency domains. Aside from learning the data's accurate spectral representation, this loss's primary focus is to improve the quality of the final restored images. The loss is named Guided Frequency Loss (GFL) and incorporates three components. Every one of them works on a different representation of the image. The first is the Charbonnier component, which targets the spatial domain by adding a penalty factor to improve the learning robustness and prevent the loss from vanishing. The second is the Laplacian Pyramid component which works on a specific compression filter of the image data that encompasses both the edge information and the content description. The third is the Gradual Frequency component, which uses a newly designed algorithm to guide the model to learn the image's frequency representation, especially the high-frequency range, where it usually struggles. 

 The main contribution of this study can be summarized below: 
\begin{itemize}
\item Introducing the GFL loss for Image Restoration. 
\item Introducing a new algorithm for learning the spectral representation of the image by gradually focusing on the hard frequencies.
\item Proving that GFL improved the PSNR metric in most of the implemented experiments.
\item Proving that the GFL improved the training of the Super Resolution in both cases of SwinIR and SRGAN.
\item Proving that the GFL performance is increased more on the constrained data.
\end{itemize}
\section{Related works}
The spectral domain gap between the real and the restored images is expressed in the theory of the spectral bias \cite{rahaman2019spectral}  and the Frequency Principle \cite{xu2019frequency} expressed above. However, profoundly describing and formulating a problem does not mean solving it. Proposing methods to solve this issue was a little bit delayed. Among the first proposals, Durall et al. \cite{durall2020watch} suggested adding a spectral regularization component to the generator loss function of the Generative Adversarial Networks (GAN) architecture. This component helped align the spectral distribution of the generated data with the original training data. This component is calculated as the binary cross entropy between the azimuthal integration component over radial frequencies of the generated output versus the mean value of the authentic data samples. The method helped to visually make the one-dimensional power spectrum of the generated data very similar to the real data. It helped to make the training of GAN more stable, the Fréchet Inception Distance (FID) is improved, and the detectability of deepfake data is becoming more difficult. However, the method was made only for GAN and general image generation without focusing on Image Restoration. Besides, it does not have tangible metrics to measure efficiency tangibly. Also, Cai et al. \cite{cai2021frequency} proposed a Frequency Domain Image Translation Framework, abbreviated as FDIT. This framework decomposes the input image into two distinct components: low frequency and high frequency. The training is done separately on these two distinct components. The two components are generated for an input image and then unified to estimate the final restored image. A similar approach has been done by Liu et al. \cite{liu2022decoupled} but targeted the  Deblurring task. They conceived a Decoupled Frequency Scheme, abbreviated as DFL, that trains two versions of the input image separately: low-pass filtered and high-pass-filtered. After that, we couple them to generate a better-restored image in qualitative and perceptual measures. Besides, Farshad et al. \cite{farshad2022net} transformed the traditional Encoder-Decoder network U-Net\cite{ronneberger2015u} into a frequency-aware architecture. They named it Y-Net and made the Encoder work in parallel as two branches, one for the spatial features and the other for the spectral features. Then, both bottlenecks are combined as input to one unified Decoder. This architecture outperformed U-Net on some benchmarks of Image Segmentation. In another proposal, Gal et al. \cite{gal2021swagan} designed a Style and Wavelet based GAN, abbreviated as SWAGAN, which makes progressive generation in the spectral representation of the image by including wavelets inside the generator and the discriminator. By integrating it inside StyleGAN2, the model generated better high-frequency content. Besides, Jung et al.\cite{jung2021spectral} designed a spectral discriminator to be integrated into the GAN architecture. It works alongside the main discriminator to focus on learning the spectral content inside the image. In another work, Fritsche et al. \cite{fritsche2019frequency} targeted the generation of low-resolution pairs of high-resolution images while building the dataset. They estimated that the interpolation-based methods are also biased against the high frequencies and alter them during the down-sampling operations. They designed for this the DSGAN model to make the downscaled images more realistic than those generated by interpolation. Another work done by Zhou et al. \cite{zhou2020guided} tried to solve the same problem by including a domain transformation stage before the Super Resolution stage to keep the details that were usually hidden in the downscaling. They also constructed the Edge Loss to enhance the quality of the texture details during the training. 
In another work, Jiang et al.\cite{jiang2021focal} designed the Focal Frequency Loss (FFL) that helps the model to focus adaptively on the hard frequencies by weighting the spectral map during the loss calculation. The weighting algorithm increases the weights assigned to the hard-to-learn frequencies while down-weighting the easy-to-learn frequencies. To our knowledge, the FFL is currently the only work that designed a new loss function for Image Restoration based on the Frequency analysis of the input image. Therefore, our study aims to find a better approach for integrating the spectral information inside the loss, specifically for Image Restoration. 
\section{Guided Frequency Loss (GFL)}
The presented GFL loss function in this study relies on the analysis of the frequency representation of the image. Therefore, this section will begin by describing the characteristics of this representation. Then, it will introduce the Laplacian Pyramid, the Charbonnier loss, the GFL global formula, and the Gradual Frequency component.  
\subsection{The Frequency Representation of the Image}
The representation of an image in the frequency domain is done using the Fourier transform. It converts the spatial representation of an image into a 2D array of complex values. Every value in this array refers to a specific frequency inside the image, figuring both the amplitude and the phase of this constituent frequency. The low frequencies represent the image's global brightness and contrast, whereas the high frequencies represent the fine details and textures in the image. 

As the image representation is in discrete and non-continuous values, the Discrete Fourier Transform (DFT) converts an image into the frequency domain by representing the constituent frequencies as complex exponential waves. The DFT can be applied to the grayscale version of the image or applied on the different channels separately. If we have an image of size \(M \times N\), and \((M,N)\) the coordinates of a specific pixel inside this image, the DFT of this image into the frequency domain is expressed by the following formulas: 

\begin{equation}
F(u,v) = \sum_{x = 0}^{M - 1}{\sum_{y = 0}^{N - 1}{f(x,y).e^{- i2\pi(\frac{ux}{M} + \frac{vy}{N})}}\ }
\end{equation}
where \(f(x,y)\) refers to the pixel value at the coordinates \((x,y)\), \((u,v)\) are the coordinates of one frequency inside the frequency domain, \(F(u,v)\) is the complex representation of the frequency value at these coordinates, \(e\) is the Euler's number, and \(i\) is the imaginary unit. The exponential component can be converted using Euler's formula below:
\begin{equation}
 e^{i\theta} = \cos\theta + i\sin\theta
 \end{equation}
Hence, the exponential component in Eq.1 can be converted into the
following formulation:
\begin{equation}
e^{- i2\pi(\frac{ux}{M} + \frac{vy}{N})} = \cos{2\pi\left( \frac{ux}{M} + \frac{vy}{N} \right)} - i\sin{2\pi\left( \frac{ux}{M} + \frac{vy}{N} \right)}
 \end{equation}
In a more intuitive explanation, the \((u,v)\) coordinates in the frequency domain refer to the angled direction of the frequency, while \(F(u,v)\) can be considered as the response to this frequency in the image. Eq. (3) shows that the frequency decomposition of the image has a real cosine part and an imaginary sine part. Both are orthogonal and constitute the vertical and the horizontal frequencies inside the image. The spatial frequency corresponds to the 2D sine components in the image.
\begin{figure}[!h]
  \centering

    \includegraphics[width=5in]{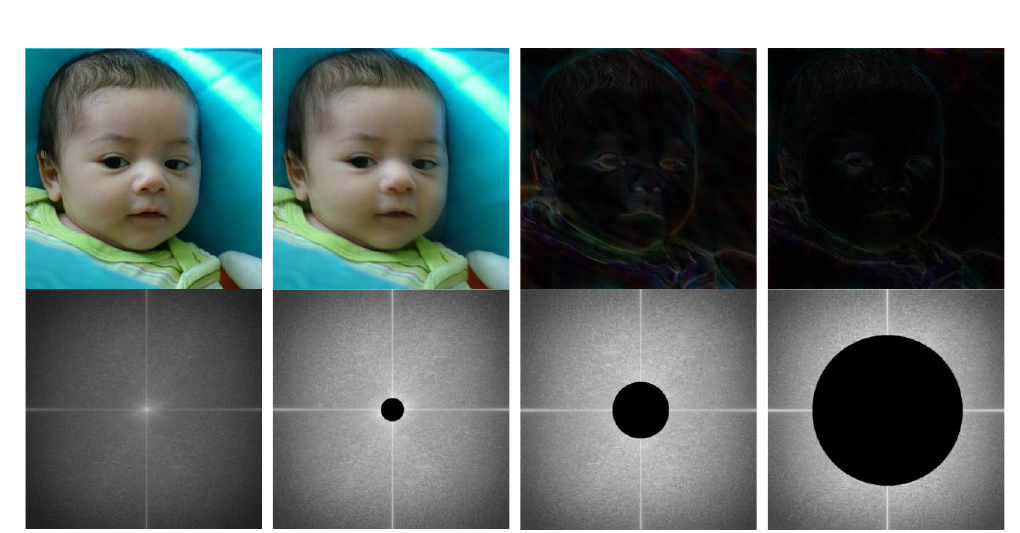}
  \caption{The effect of conserving different levels of frequencies on the visual quality of the image.}
  \label{fig_1}
\end{figure}

In Fig. \ref{fig_1}, the effect of the high frequencies on the image is visually explained. On the top side of the Figure, the spatial representation of the image is displayed, while the frequency representation is displayed on the bottom. In the first column, the original image is displayed without any alteration. The entire frequency spectrum exists. Even by zooming inside the image, we can see that most global and fine detail are shown approaching it from the authentic representation of the captured scene. In the second column, a high pass filter is applied to the image, eliminating some small range of frequencies. We see that the image is visually altered and far from the scene\textquotesingle s authentic representation. Similarly, a larger range of frequencies is cropped out for the third and fourth columns. We can see that most visual content is located in the low-frequency spectrum of the image. The higher the frequency, the more difficult to perceive its influence on the image.

In the fourth column, a large part of the high frequencies is kept in the image, although their visual signification may not be well perceived in the spatial domain. However, these fine details represent a necessary part of the image to make it more similar to the captured scene. Thus, learning these high frequencies should be explicitly incorporated in any image restoration model to bridge the gap between the restored image and its authentic representation. In addition, learning the complete map of the frequencies represents a measure of the efficiency of the restoration model.

\subsection{Laplacian Pyramid}
The Laplacian Pyramid \cite{burt1987laplacian} is a filtering algorithm that helps to grasp important frequency information in the image, such as the edge, the objects, and the textures. It is widely used in image compression, texture analysis, and image content retrieval \cite{paris2011local}.  Its name came from the process of down-sampling the image by order of one octave (division by 2). We apply the Gaussian filter at every step of the down-sampling. The Laplacian Pyramid is established first by generating the difference between this Gaussian-blurred image and its up-sampled counterpart. This will be named the Laplacian Image and contains the high-frequency details hidden in the next lower level of the Pyramid. Next, the Laplacian image is down-sampled and added to the next level inside this Pyramid. The number of sequences is named the depth level of the Pyramid. The underlining concept behind it is to progressively collect the high-frequency details of the image at different levels of resolutions.

Let \(I\) be the input  image of square size \(j \times j\). Let \(d()\) be the operation that applies a Gaussian blurring filter and decimates into a new image \(d(I)\) of size \(\frac{j}{2} \times \frac{j}{2}\). On the reverse, let \(u()\) be the up-sampling operation that applies a smoothing filter and then extend the image size to its twice, resulting in a new image \(u()\) of size \(2j \times 2j\).

We begin first by constructing the Gaussian Pyramid \(G(I) = \lbrack I_{0},\ I_{1},\ldots,\ I_{N}\rbrack\), where \(I_{0}\) corresponds to the input image \(I\), \(I_{n}\)is the repeated application of the down-sampling operation \(d()\) to \(I_{0}\). \(N\) is the count of down-sampling levels in the Pyramid. The lowest level should be of a size no less than \(8 \times 8\) pixels to conserve minimal spatial details inside the image.

\begin{figure}[!h]
  \centering
    \includegraphics[width=5in]{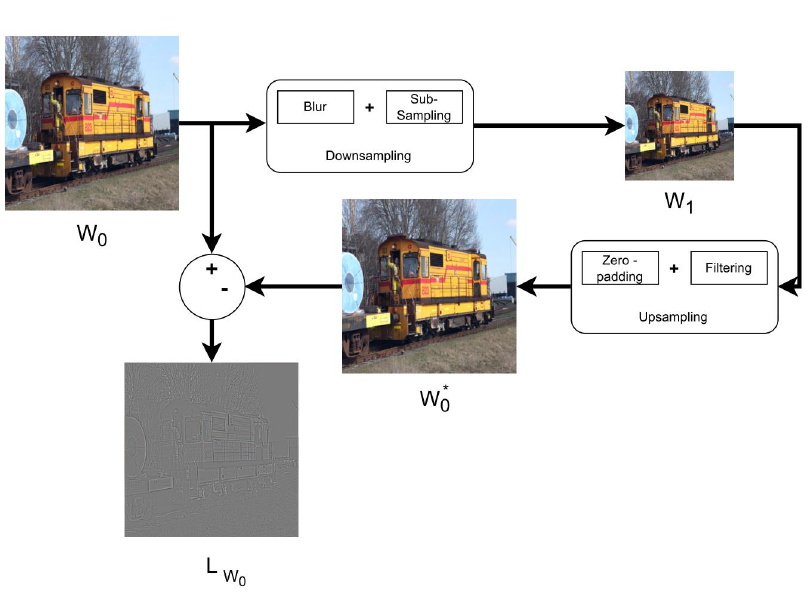}
  \caption{Generation of the Laplacian Pyramid of depth 1 from an input Image \(W_{0}\)}
  \label{fig_2}
\end{figure}

The Laplacian Pyramid \(L(I) = \lbrack h_{0},h_{1},\ldots,h_{N}\rbrack\) is built by repeatedly calculating the coefficients \(h_{n}\) at every level. \(h_{n}\) takes two adjacent levels from the Gaussian Pyramid, up-samples the smallest in size, and then calculates the difference between them. The procedure is expressed in the following Equation:
\begin{equation}
    h_{n} = I_{n} - u\left( I_{n + 1} \right)
\end{equation}
For the final coefficient level \(h_{N}\), it is equal to the final level of the Gaussian Pyramid \(I_{N}\), since it does not have a lower level. Each level \(h_{n}\) is intended to extract the image's main structure at a specific resolution. The Laplacian Pyramid is an invertible process, where we can construct the original image \(I_{0}\) by following a backward process starting from \(h_{N} = \ I_{N}\), and generating the upper level of the Gaussian Pyramid by using the Equation:
\begin{equation}
I_{n} = u\left( I_{n + 1} \right) + h_{n}
 \end{equation}
By recursively repeating the above operation, we will construct at the end the original image \(I_{0}\).
In Fig. \ref{fig_2}, a Laplacian Pyramid of depth one is applied to an image. 

This example shows that the Laplacian Pyramid outperforms ordinary filters in summarizing the image\textquotesingle s main features, especially the objects\textquotesingle{} edges, and structures. This explains why it was mainly conceived for image encoding and compression, as it is a better summarizing filter for the image\textquotesingle s content.

\begin{figure}[!h]
  \centering
    \includegraphics[width=2.5in]{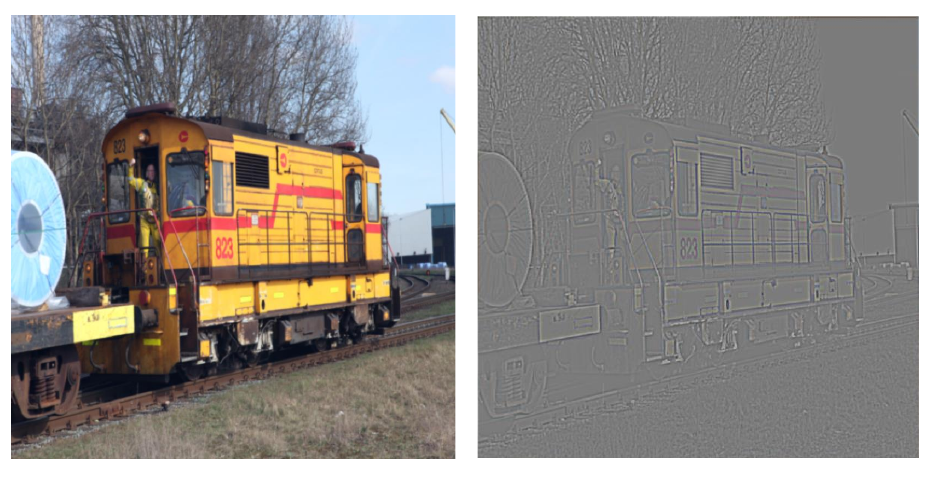}
  \caption{The image \(W_{0}\) and its Laplacian Pyramid of depth 1}
  \label{fig_3}
\end{figure}

A closer look at the Image \(W_{0}\) and its Laplacian filter of depth one is displayed in Fig. \ref{fig_3}. We can see that this filter does extract not only the edge information from the image, but also the color and the visual characteristics of the objects.

In this study, the Laplacian Pyramid of depth one is considered. Concatenating information from the other levels of the Pyramid has not been investigated.


\subsection{Charbonnier Loss}
The Charbonnier loss \cite{charbonnier1994two} is a differentiable variant of the \(L_{1}\) loss (absolute difference). It adds a very small constant \(\varepsilon\) to ensure more robustness during the model\textquotesingle s training. It was proven efficient in image restoration \cite{lai2017deep}. It acts by using the Charbonnier penalty function defined below:
\begin{equation}
    \rho(x) = \sqrt{x^{2} + \varepsilon^{2}}
\end{equation}

when applied to the Image Restoration tasks, the Charbonnier loss will be expressed by the following Equation:
\begin{equation}
 L\left( I_{R},\ I_{HQ} \right) = \sqrt{\left\| I_{R} - I_{HQ} \right\|^{2} + \varepsilon^{2}}
\end{equation}
where \(I_{R}\) is the restored image predicted by the model during the training, \(I_{HQ}\) is the high-quality image serving as the targeted ground truth. The constant \(\varepsilon\) is set empirically to \(10^{- 3}\) during the experiment. It smoothens the loss while converging to 0 and prevents it from vanishing during the training.

\subsection{The Guided Frequency Loss (GFL) formula}
The proposed loss in this paper relies upon three main concepts. The first concept is the frequency representation of the image and the explicit need to guide the Image Restoration to learn the high-frequency components artificially. The second concept is the Laplacian Pyramid and its strength in modeling the main structures inside the image. The third concept is the Charbonnier penalty function which enhances the robustness during the training. The global expression of the loss function is summarized in the following Equation:
\begin{equation}
    GFL\ \left( I_{R},\ I_{HQ} \right) = \sqrt{{Ch}_{C} + {\Pi_{C}\  + \Theta}_{C}}
\end{equation}

where \({Ch}_{C}\) is the Charbonnier component, \(\Pi_{C}\) is the Laplacian Pyramid component, and \(\Theta_{C}\) is the Gradual Frequency component.

The Charbonnier component \({Ch}_{C}\) is expressed by the following Equation and serves to compare by the intermediate of the Charbonnier penalty the squared absolute difference between the predicted image and the ground truth in the spatial domain. The square power is used to compensate the global square root of the global expression. If we consider \(I_{R}\) as the restored image predicted by the model, and \(I_{HQ}\) as the high-quality image or the ground truth, \({Ch}_{C}\) is expressed by the following Equation:
\begin{equation}
    {Ch}_{C} = \left\| I_{R} - I_{HQ} \right\|^{2}{+ \varepsilon}^{2}
\end{equation}

The Laplacian Pyramid component \(\Pi_{C}\) is the squared absolute difference between the Laplacian Pyramid of depth 1 of the predicted image \(\Pi\left( I_{R} \right)\) and the Laplacian Pyramid of depth 1 of the ground truth image \(\Pi(I_{HQ})\), it is described by the following Equation:
\begin{equation}
    \Pi_{C} = \ \left\| \Pi\left( I_{R} \right) - \Pi(I_{HQ}) \right\|^{2}
\end{equation}

Finally, the Gradual Frequency \(\Theta_{C}\) component is expressed by the following Equation:
\begin{equation}
\Theta_{C} = \left\| \Theta\left( I_{R} \right) - \Theta\left( I_{HQ} \right) \right\|^{2}
\end{equation}

where \(\Theta\) is the Frequency High-Pass filter applied at every epoch of the training. This component gradually increases the band of the high frequencies being learned. The band considered at every stage of the training is to be tuned depending on the number of epochs and the model itself based on experimental observations. However, as a general rule, we need to consider only very high frequencies at the earlier stages of the training, then increment gradually another band of high-frequencies until we include at the last stage of the training all the high-frequencies that are difficult to learn by the model. As a result, \(\Theta_{C}\) will calculate the squared absolute difference between the filtered version of the predicted image
\(\Theta\left( I_{R} \right)\) and the filtered image of the ground truth \(\Theta\left( I_{HQ} \right)\).
In Algorithm \ref{Algorithm_1}, the process for the band allocation for \(\Theta\) during the training is summarized. The algorithm has seven inputs to be set experimentally based on the model, the dataset, the task, the number of epochs, and the experimental observations to be considered.

\begin{algorithm}[http]
  \caption{The frequency band allocation for $\Theta_C$}
  \label{Algorithm_1}
  \SetKwInOut{Input}{Input}
  \SetKwInOut{Output}{Output}
  
  \Input{%
\begin{enumerate}
\def\labelenumi{\arabic{enumi}.}
\item
  \(\omega_{0}\): The initial frequency threshold
\item
  \(\omega_{F}\): The final frequency threshold
\item
  \(N\): The number of the training epochs
\item
  \(S:\)The number of stages in \(\Theta_{C}\)
\item
  \(HPF(\omega_{i})\): High-Pass filter of frequencies
  \({> \omega}_{i}\)
\item
  \(b_{alloc}\): Band allocation mode ( \(static\ \)or \(dynamic\))
\item
  \(L_{t}\) : Loss threshold for the dynamic band allocation
\end{enumerate}
  }
  
  $\Theta \leftarrow HPF(\omega_0)$\;
  $\omega_{\text{prev}} = \omega_0$\;
  
  \For{$i$ \textbf{in} $(1..N)$}{
    // Inside every epoch of the training\;
    
    \If{$(b_{\text{alloc}} = \text{static}) \textbf{and} (i \mod S = 0) \textbf{or} (b_{\text{alloc}} = \text{dynamic}) \textbf{and} (\text{GFL} < L_t)$}
    {
      // Pass to the next stage\;
      $\omega_{\text{prev}} = \omega_{\text{prev}} - (\omega_0 - \omega_F) \div S$\;
      
      // Check if we are at the last stage\;
      \If{$(\omega_{\text{prev}} - \omega_F) < ((\omega_0 - \omega_F) \div S)$}{
        $\omega_{\text{prev}} = \omega_F$\;
      }
      
      $\Theta \leftarrow HPF(\omega_{\text{prev}})$\;
      Update $GFL$ loss with the new value of $\Theta$\;
      
      \If{$\omega_{\text{prev}} = \omega_F$}{
        $\Theta$ is fixed for the next epochs\;
        \textbf{break}\;
      }
    }
  }
\Output{$\Theta$}
\end{algorithm}

\(\Theta\) acts on the image as a High-Pass filter that progressively enlarges its band during training. It starts from an initial frequency threshold \(\omega_{0}\) at the beginning of the training. At the end of the training, it includes all the frequencies between \(\omega_{0}\) and the final frequency threshold \(\omega_{F}\). It has two possible modes: static and dynamic. The mode value is saved in the variable \(b_{alloc}\). In the static mode, the user sets a number of stages \(S\ \)for incrementing the frequency band. The algorithm divides the number of epochs \(N\) by the number of stages \(S\), and also divides the range of frequencies between \(\omega_{0}\) and \(\omega_{F}\) by \(S\). For every stage in training, \(\Theta\) includes the following range of frequencies. This mode does not affect the training progress.
Concerning the dynamic mode of the algorithm, it depends on the \(GFL\) loss progress and not on the number of epochs. It tests if \(GFL\) loss is less than a fixed threshold to ensure that the model has learned well the previous band of frequencies. In every epoch, it checks if the \(GFL\) has fallen below the threshold \(L_{t}\) to increment the next stage of frequencies to \(\Theta\). The stages of frequencies are
divided based on the value set in \(S\). If we put a low threshold, the training progress risk falling into one stage without being able to pass to the next one.
Concerning the components of the \emph{GFL} loss, \({Ch}_{C}\) and \(\Pi_{C}\) do not change. However, the utility of \(\Theta_{C}\) may not be perceived until the right tuning of the parameters is made based on the experimental observations.
\section{Experiments}
This section assesses the validity of the \(GFL\) loss for different use cases of Image Restoration. It begins by explaining the experimental configurations. Then, it analyses the different results obtained.
\subsection{Experimental configurations}
This study considered three main factors to estimate the performance of the \(GFL\) loss. In the following subsections, we will describe these three factors and cite the options selected for every factor.
\subsubsection{Image Restoration tasks}
Due to the high number of Image Restoration tasks, we chose from them the most popular: the Image Super Resolution and the Image Denoising tasks. We did similar to the popular SwinIR \cite{liang2021swinir} architecture. They introduced the architecture for Image Restoration but tested it on three tasks only: Super Resolution, Image Denoising, and JPEG Compression Artifact Reduction. In fact, if a new method succeeds in these domains, it will probably succeed in most Image Restoration tasks.
\subsubsection{Models used in Image Restoration}
Different models are used in Image Restoration. However, most of them
belong to one of these three types of architectural patterns: Generative
Adversarial Networks \cite{goodfellow2020generative}, Encoder-Decoder models \cite{liang2021swinir}, \cite{conde2023swin2sr}, \cite{ronneberger2015u}, and Diffusion Models \cite{ho2020denoising}, \cite{rombach2022high}. Hence, this study
selected one state-of-the-art model from each category. For the GAN
category, it selects SRGAN \cite{ledig2017photo} for Super Resolution. For Denoising,
it selects Denoising-GAN (inspired by SRGAN \cite{ledig2017photo} and adapted for
Denoising by removing the up-sampling step at the end). We selected SwinIR \cite{liang2021swinir} for the Encoder-Decoder category for both Super Resolution and Denoising as they have variants. For the third category of Diffusion models, the Iterative Diffusion Model (SR3) \cite{saharia2022image} was also selected for both Super Resolution and Denoising. The same architecture is kept for both tasks. For the GAN model, the \emph{GFL} loss is included during the training of the generator as it cannot be included in the discriminator.
\subsubsection{Constrained/ Unconstrained Datasets}
Datasets for image restoration can be generic or constrained. Generic means that they are unconstrained, i.e., not having solid constraints about the class of the objects. Therefore, to test the GFL loss on both types of datasets, this study took three datasets:
\begin{enumerate}
\def\labelenumi{\arabic{enumi}.}
\item
  \textbf{A Generic dataset:} we select the DIV2K \cite{timofte2017ntire}, \cite{agustsson2017ntire} dataset in training. It is divided into 800 images in the training and 100 in the validation. We selected BSD100 \cite{martin2001database} for the test, which is exactly 100 images. All of them were used for the test. This method was used in many Image Restoration benchmarks \cite{liang2021swinir}, \cite{conde2023swin2sr}.
\item
  \textbf{Constrained dataset 1 (Faces):} we randomly selected 544 images from CelebA-HQ \cite{liu2015deep} dataset. 480 images were reserved for training, 20 for validation, and 44 for the test. This dataset is constrained only on human faces only. Therefore, all images are depictions of human faces.
\item
  \textbf{Constrained dataset 2 (Vehicles License Plates: VLP):} we randomly selected 593 images of Vehicle License plates. From them, 549 are reserved for training, and 44 are reserved for the test. The license coding pattern is kept the same for all the images. However, the samples differ very much regarding the licensing code, view, and image quality. This dataset is more constrained that the Faces dataset. These two datasets help to estimate the performance of the
  \emph{GFL} loss when we get more constraints on the data.
\end{enumerate} For the Super Resolution task, we considered the scale \(\times 4\). The low-resolution pairs are obtained by down-scaling the high-resolution images using a bicubic interpolation. For the Denoising task, the noisy pairs are obtained by adding an artificial Gaussian noise ( \(\sigma = 0.15,\ \mu = 0\ \)) to the images.
\subsubsection{Implementation details}
For every used model, the code is trained on a Lambda AI server having: 8 GPUs NVIDIA QUADRO 8000 (48 GB GDDR6), with two processors Intel Xeon Silver 4216 having 16 cores, and 512 GB of RAM. Every model is trained from scratch on the chosen loss function without pretraining on extra data. We fixed the number of epochs for all the experiments to 100 epochs and the number of batch size to 8. The component \(\Theta_{C}\) may variate from one experiment to another according to Algorithm \ref{Algorithm_1}.
\subsection{Results and Analysis}
\subsubsection{GFL performance on the Super Resolution task}
The impact of the \emph{GFL} loss on the training of the Super
Resolution models is made in Table \ref{table_1}. The best performance of the \emph{GFL} loss is shown in two cases. The first is the training of the SwinIR model (both generic and constrained datasets).

\begin{table}[htbp]
\small
  \caption{Comparison of \emph{GFL} loss for the Super Resolution and the Denoising tasks on three different models and three different datasets, the \textcolor{blue}{blue} color is for the best result, the \textcolor{red}{red} color is for the second-best results.}
  \label{table_1}
  \centering
  \begin{tabular}{lllrrll}
    \toprule
     &  & & \multicolumn{2}{c}{Super-Resolution} & \multicolumn{2}{c}{Denoising}    \\
     \cmidrule(l){4-7}
    Model    & Dataset  & Loss & PSNR & SSIM & PSNR & SSIM  \\
    \midrule
    \multirow{15}{8em}{SwinIR}   & \multirow{5}{5em}{Generic}  & MSE  & 24.497  & 0.7428 & 24.165  & \textcolor{blue}{0.6891}   \\
             &          & Charbonnier   & \textcolor{red}{24.507}  & \textcolor{red}{0.7482} & \textcolor{blue}{24.276}  & \textcolor{red}{0.6881}    \\
             &          & Edge  & 20.018  & 0.6984 & 23.473  & 0.6435    \\
             &          & FFL  & 24.349  & 0.7384 & 23.908  & 0.6601   \\
             &          & Our  & \textcolor{blue}{24.541}  & \textcolor{blue}{0.7617} & \textcolor{red}{24.188}  & 0.6847  \\
    \cmidrule(r){2-7}             
                        & \multirow{5}{1em}{Faces}  & MSE  & 29.289  & \textcolor{red}{0.8945} & 25.588  & 0.7672   \\
             &          & Charbonnier   & \textcolor{red}{29.446}  & 0.8912  & \textcolor{blue}{26.119}  & \textcolor{red}{0.7901}   \\
             &          & Edge  & 20.651  & 0.7849 & 25.171  & 0.7716   \\
             &          & FFL  & 28.523  & 0.8792 & 23.451  & 0.6603   \\
             &          & Our  & \textcolor{blue}{29.723}  & \textcolor{blue}{0.8979} & \textcolor{red}{25.945}  & \textcolor{blue}{0.8979}   \\    
    \cmidrule(r){2-7}
                               & \multirow{5}{1em}{VLP}  & MSE & \textcolor{red}{24.770} & 0.8928 & \textcolor{red}{25.048} & \textcolor{red}{0.8922}  \\
             &          & Charbonnier   & 24.375  & \textcolor{red}{0.8944} & 24.416  & 0.8838   \\
             &          & Edge  & 18.886  & 0.7455 & 22.438  & 22.438   \\
             &          & FFL  & 23.061  & 0.8464 & 23.617  & 0.8473   \\
             &          & Our  & \textcolor{blue}{26.698}  & \textcolor{blue}{0.9310} & \textcolor{blue}{25.305}  & \textcolor{blue}{0.9004}   \\
             
    \cmidrule(r){1-7}    
    
    \multirow{15}{1em}{SRGAN}   & \multirow{5}{1em}{Generic}  & MSE  & \textcolor{blue}{17.391}  & \textcolor{blue}{0.6882} & \textcolor{red}{22.655}  & \textcolor{red}{0.8234}   \\
             &          & Charbonnier   & 16.697  & 0.6576 & 22.631  & 0.8249   \\
             &          & Edge  & 6.056  & 0.2802  & 16.370  & 0.7366   \\
             &          & FFL  & 14.268  & 0.5606  & 17.080  & 0.6599  \\
             &          & Our  & \textcolor{red}{17.066}  & \textcolor{red}{0.6712}  & \textcolor{blue}{23.084}  & \textcolor{blue}{0.8397}  \\
    \cmidrule(r){2-7}             
                               & \multirow{5}{1em}{Faces}  & MSE  & 20.333  & 0.8176 & \textcolor{red}{24.524}  & 0.8806   \\
             &          & Charbonnier   & \textcolor{red}{20.476}  & \textcolor{red}{0.8244} & \textcolor{blue}{25.322}  & \textcolor{blue}{0.8928}    \\
             &          & Edge  & 14.033  & 0.6815 & 17.247  & 0.7743    \\
             &          & FFL  & 15.527  & 0.6921 & 17.267  & 0.7381    \\
             &          & Our  & \textcolor{blue}{20.534}  & \textcolor{blue}{0.8270} & 24.351  & \textcolor{red}{0.8814}    \\    
    \cmidrule(r){2-7}
                               & \multirow{5}{1em}{VLP}  & MSE  & \textcolor{red}{18.884}  & \textcolor{red}{0.8197} & \textcolor{red}{24.086}  & 0.8932    \\
             &          & Charbonnier   & 17.290  & 0.7831 & 23.923  & 0.8973    \\
             &          & Edge  & 10.732  & 0.6601 & 13.501  & 0.7505   \\
             &          & FFL  & 12.908  & 0.5607 & 14.811  & 0.6764   \\
             &          & Our  & \textcolor{blue}{22.379}  & \textcolor{blue}{0.8945} & \textcolor{blue}{25.010}  & \textcolor{blue}{0.9163}   \\
    \cmidrule(r){1-7}
    \multirow{10}{1em}{SR3}   & \multirow{5}{1em}{Generic}  & MSE  & 15.119  & \textcolor{blue}{0.4535} & \textcolor{red}{15.266}  & 0.35777   \\
             &          & Charbonnier   & \textcolor{blue}{17.403}  & 0.3769  & \textcolor{blue}{15.799}  & \textcolor{red}{0.27557}   \\
             &          & Edge  & 12.199  & 0.2581  & 12.285  & 0.27044   \\
             &          & FFL  & \multicolumn{1}{l}{\textbf{-}}  & \multicolumn{1}{l}{\textbf{-}} & \multicolumn{1}{l}{\textbf{-}}  & \multicolumn{1}{l}{\textbf{-}}    \\
             &          & Our  & \textcolor{red}{16.525}  & \textcolor{red}{0.3856}  & 15.108  & \textcolor{blue}{0.28936}  \\
    \cmidrule(r){2-7}             
                               & \multirow{5}{1em}{Faces}  & MSE  & \textcolor{red}{19.876}  & \textcolor{blue}{0.6296} & 12.814  & \textcolor{blue}{0.48021}   \\
             &          & Charbonnier   & 16.534  & 0.4093 & 12.922  & 0.47772    \\
             &          & Edge  & 10.161  & 0.1305 & 12.069  & 0.33575    \\
             &          & FFL  & \multicolumn{1}{l}{\textbf{-}}  & \multicolumn{1}{l}{\textbf{-}} & \multicolumn{1}{l}{\textbf{-}}  & \multicolumn{1}{l}{\textbf{-}}    \\
             &          & Our  & \textcolor{blue}{20.088}  & \textcolor{red}{0.6252} & \textcolor{blue}{17.501}  & 0.45653   \\      

    \bottomrule
  \end{tabular}
\end{table}

 The second case is the training of the GAN models for the constrained datasets (Faces and VLP). In these two cases, it outperforms other losses. We see also that GFL\textquotesingle s best performance is noted on the VLP dataset, then the Faces dataset, then the Generic dataset.
\subsubsection{GFL performance on the Denoising task} 
The impact of the \emph{GFL} loss on the training of the denoising models is made in Table \ref{table_1}. The best performance of the \emph{GFL} loss is shown during the training of the GAN models in generic datasets. For the other cases, we can conclude that the performance of \emph{GFL} increases on constrained datasets, especially VLP, which is more constrained than the Faces dataset. The \emph{GFL} performance can be improved more by tuning Algorithm \ref{Algorithm_1}.

\subsection{Ablation Study} 
In Table \ref{table_Ablation_Study}, we assess the increment made by the components \(\Pi_{C}\), \(Ch_{C}\), and \(\Theta_{C}\), separately. The table shows the performance during the training of the SwinIR \cite{liang2021swinir} for the super-resolution task on two datasets: the Generic dataset and the VLP dataset, described above. The results demonstrate that every one of the components increments the efficiency of the training by a significant margin. 

\begin{table}[htbp]
  \caption{Ablation study of \(\Pi_{C}\), \(Ch_{C}\), and \(\Theta_{C}\) in the \emph{GFL} loss}
  \label{table_Ablation_Study}
  \centering
  \begin{tabular}{lllll}
    \toprule
     & \multicolumn{2}{c}{Generic} & \multicolumn{2}{c}{VLP}    \\
     \cmidrule(l){2-5}
    Components & PSNR & SSIM & PSNR & SSIM \\
    \midrule
    $\sqrt{\Pi_{C}}$  & 20.018 & 0.6984 & 18.886 & 0.7455  \\
    $\sqrt{\Pi_{C} + Ch_{C}}$ & 24.343 & 0.7364 & 23.805 & 0.8725  \\
    $\sqrt{\Pi_{C} + Ch_{C} + \Theta_{C}}$ & 24.541 & 0.7617 & 26.698  & 0.9310  \\
    \bottomrule
  \end{tabular}
\end{table}

The parameters used for \(\Theta_C\) in these experiments are: \(\omega_0 = 255\), \(\omega_F = 10\), \(N = 100\), \(S = 2\), \(b_{alloc} = static\). 
\subsection{Discussion} 
To get more into the details, we made an average over all the implemented experiments to judge the efficiency of the \emph{GFL} loss regarding nine different factors. We made two separate tables for the PSNR and the SSIM metrics. The results are shown in Table \ref{table_3} and Table \ref{table_4}.

\begin{table}[htbp]
  \caption{Comparison of average PSNR considering different factors, the \textcolor{blue}{blue} color is for the best result, and the \textcolor{red}{red} is for the second-best results.}
  \label{table_3}
  \centering
  \begin{tabular}{llllllllll}
    \toprule
    Loss    & Total  & SR & Denoising & SwinIR & GAN & SR3 & Generic & Gaces & VLP  \\
    \midrule
    MSE & \textcolor{red}{21.519}  & \textcolor{red}{21.270} & 21.768 & \textcolor{red}{25.560} & \textcolor{red}{21.312} & \textcolor{red}{15.769} & 19.849 & \textcolor{red}{22.071} & \textcolor{red}{23.197}  \\
     Char & 21.383 & 20.841 & \textcolor{red}{21.926} & 25.523 & 21.057 & 15.665 & \textcolor{blue}{20.219} & 21.803 & 22.501   \\
     Edge & 15.956 & 14.092 & 17.819 & 21.773 & 12.990 & 11.679 & 15.067 & 16.555 & 16389       \\
     FFL & 19.898 & 19.773 & 20.023 & 24.485 & 15.310 & - & 19.902 & 21.193 & 18.599    \\
     Our &  \textcolor{blue}{22.378} & \textcolor{blue}{22.194} & \textcolor{blue}{22.562} & \textcolor{blue}{26.067} & \textcolor{blue}{22.071} & \textcolor{blue}{17.306} & \textcolor{red}{20.085} & \textcolor{blue}{23.024} & \textcolor{blue}{24.848}  \\
    \bottomrule
  \end{tabular}
\end{table}             

\begin{table}[!h]
  \caption{Comparison of average SSIM considering different factors, the \textcolor{blue}{blue} color is for the best result, and the \textcolor{red}{red} is for the second-best results.}
  \label{table_4}
  \centering
  \begin{tabular}{llllllllll}
    \toprule
    Loss    & Total  & SR & Denoising & SwinIR & GAN & SR3 & Generic & Gaces & VLP  \\
    \midrule
    MSE & \textcolor{red}{0.7327}  & \textcolor{red}{0.7423} & \textcolor{blue}{0.7230} & 0.8131 & \textcolor{red}{0.8205} & \textcolor{blue}{0.4803} & \textcolor{blue}{0.6258} & \textcolor{red}{0.4120} & \textcolor{red}{0.8745}  \\
     Char & 0.7072 & 0.6981 & 0.7163 & \textcolor{red}{0.8160} & 0.8133 & 0.3849 & 0.5952 & 0.7142 & 0.8646   \\
     Edge & 0.5873 & 0.5299 & 0.6447 & 0.7531 & 0.6472 & 0.2487 & 0.4812 & 0.5797 & 0.7577       \\
     FFL & 0.7100 & 0.7129 & 0.7071 & 0.7720 & 0.6480 & - & 0.5660 & 0.7425 & 0.7327    \\
     Our &  \textcolor{blue}{0.7346} & \textcolor{blue}{0.7493} & \textcolor{red}{0.7199} & \textcolor{blue}{0.8277} & \textcolor{blue}{0.8384} & \textcolor{red}{0.4392} & \textcolor{red}{0.6054} & \textcolor{blue}{0.7465} & \textcolor{blue}{0.9106}  \\
    \bottomrule
  \end{tabular}
\end{table} 
Overall, we note that \emph{GFL} elevated the PSNR on most experiments and factors, except on the Generic dataset, where it has a small margin of 0.134 below the best-performing loss. This does not mean that GFL will fail on generic datasets, which is disproved in Table \ref{table_1}. It means that its performance in this specific case is not optimal, but if we consider its performance on the task type and the model, this sub-optimality is compensated and hidden in most cases, as shown in Table \ref{table_1}. Concerning SSIM, it was elevated by \emph{GFL} for all the factors except three cases: the Denoising task, the SR3, and the Generic datasets. Concerning the tasks, the \emph{GFL} performance is more noted on the Super Resolution task than the Denoising task, especially when using the SwinIR and SRGAN models.
Concerning the dataset type, \emph{GFL} works better on constrained datasets. GFL performance on the VLP dataset is better than the Faces dataset. The more the data is constrained, the more its high-frequency patterns will be learnable and catchable by GFL. Due to their diversity in the generic data, the frequency patterns are more difficult to learn than the spatial patterns due to their complexity, especially for the high-frequency range.
\subsection{Conclusion} 
High frequencies remain an inherent challenge in the Image Restoration domain. Mimicking authentic data requires learning this set of frequencies, even if they are difficult to perceive visually in the image. Removing them leads to the appearance of detectable or undetectable artifacts inside the image. This study aimed to target this problem by introducing a new loss function that helps the Image
Restoration model learn better from the frequency domain. It is named the Guided Frequency Loss (\emph{GFL}) and comprises three components: the Charbonnier component, the Laplacian Pyramid component, and the Gradual Frequency component. Experiments validate that the GFL loss increased the PSNR of the restored images in most cases. Also, the performance is noted, especially for the Super Resolution task when using SwinIR and SRGAN. This indicates its potential for other Encode-Decoder models or GAN-based models. Also, we noted that the more the dataset is constrained, the more the \emph{GFL} loss helps to improve the results. This can be explained intuitively that the frequency patterns are easier to learn in the constrained datasets, leading to a better quality of the restored image. Although other losses performed better than \emph{GFL} in some experiments\emph{,} its
performance may improve by fine-tuning the Gradual Frequency component. As an extension of the current study, the \emph{GFL} can be improved by making a more generic version of the Gradual Frequency component. Also, we can consider a weighting pattern for the different components or encompassing other levels in the Laplacian Pyramid. Moreover, as a next research direction, solving the spectral bias internally inside the model will be very useful. Although this problem may be challenging due to limitations in the underlying feature extractors themselves, this opens new horizons, especially for the Image Restoration domain.

\section*{Acknowledgements}
The authors thank Prince Sultan University for their support and funding in conducting this research
\bibliographystyle{elsarticle-num}

\bibliography{mybibfile}

\end{document}